\documentclass[letterpaper, 10pt, conference]{ieeeconf} 

\IEEEoverridecommandlockouts       
\usepackage{times} 
\usepackage{graphicx} 
\usepackage{bm} 
\usepackage{amsmath} 
\usepackage{amssymb} 
\usepackage{color} 
\usepackage{booktabs}
\usepackage[ruled,linesnumbered]{algorithm2e} 
\usepackage[short]{optidef} 
\usepackage{gensymb} 
\usepackage{multirow}
\usepackage[mathscr]{eucal}
\usepackage{threeparttable}
\usepackage{comment}

\usepackage{xparse,soul} 
\soulregister\cite7 
\soulregister\citep7 
\soulregister\citet7 
\soulregister\ref7 
\soulregister\pageref7 
\usepackage{makecell}

\usepackage{fancyhdr}
\fancypagestyle{firstpage}{
    \chead{This paper has been accepted for publication at the 2026 International Conference on Robotics and Automation.}}
\usepackage{cite}
\title{A Single-Fiber Optical Frequency Domain Reflectometry (OFDR)-Based Shape Sensing of Concentric Tube Steerable Drilling Robots }

\author{Yash Kulkarni$^{*1}$, Mobina Tavangarifard$^{*1}$, Daniyal Maroufi$^{1}$, Mohsen Khadem$^{2}$, Justin E. Bird $^{3}$, \\ Jeffrey H. Siewerdsen$^{4}$, and Farshid Alambeigi$^{1}$
\thanks{*These authors contributed equally to this work.}
\thanks{**This work was supported by the Collaborative Accelerator for Transformative Research Endeavors grant, jointly awarded by The University of Texas at Austin and The University of Texas MD Anderson Cancer Center.}
\thanks{$^{1}$Y.~Kulkarni, M.~Tavangarifard, D.~Maroufi and F.~Alambeigi are with the Walker Department of Mechanical Engineering and Texas Robotics at The University of Texas at Austin, Austin, TX, 78712, USA. Email: \{kulkarni.yash08, mtavangarifard\}@utexas.edu. \{farshid.alambeigi\}@austin.utexas.edu}%
\thanks{$^{2}$M.~Khadem is with the School of Informatics, University of Edinburgh, UK. }
\thanks{$^{3}$J.~E.~ Bird is with the Department of Orthopedic Oncology, Division of Surgery, The University of Texas M.D. Anderson Cancer Center, Houston, TX, USA} 
\thanks{$^{4}$J.~H.~ Siewerdsen is with the Department of Imaging Physics, Division of Diagnostic Imaging, The University of Texas MD Anderson Cancer Center, Houston, Texas, USA}}

\usepackage{color} 

\begin{document}
\maketitle
\thispagestyle{firstpage}
\pagestyle{empty}
		
\begin{abstract} 
This paper introduces a novel shape-sensing approach for Concentric Tube Steerable Drilling Robots (CT-SDRs) based on Optical Frequency Domain Reflectometry (OFDR). Unlike traditional FBG-based methods, OFDR enables continuous strain measurement along the entire fiber length with enhanced spatial resolution. In the proposed method, a Shape Sensing Assembly (SSA) is first fabricated by integrating a single OFDR fiber with a flat NiTi wire. The calibrated SSA is then routed through and housed within the internal channel of a flexible drilling instrument, which is guided by the pre-shaped NiTi tube of the CT-SDR. In this configuration, the drilling instrument serves as a protective sheath for the SSA during drilling, eliminating the need for integration or adhesion to the instrument surface that is typical of conventional optical sensor approaches.
The performance of the proposed SSA, integrated within the cannulated CT-SDR, was thoroughly  evaluated under free-bending conditions and during drilling along multiple J-shaped trajectories in synthetic Sawbones phantoms. Results demonstrate accurate and reliable shape-sensing capability, confirming the feasibility and robustness of this integration strategy.
\end{abstract}

\section{Introduction}
\begin{figure}[t] 
    \centering
    \includegraphics[width=0.9\linewidth]{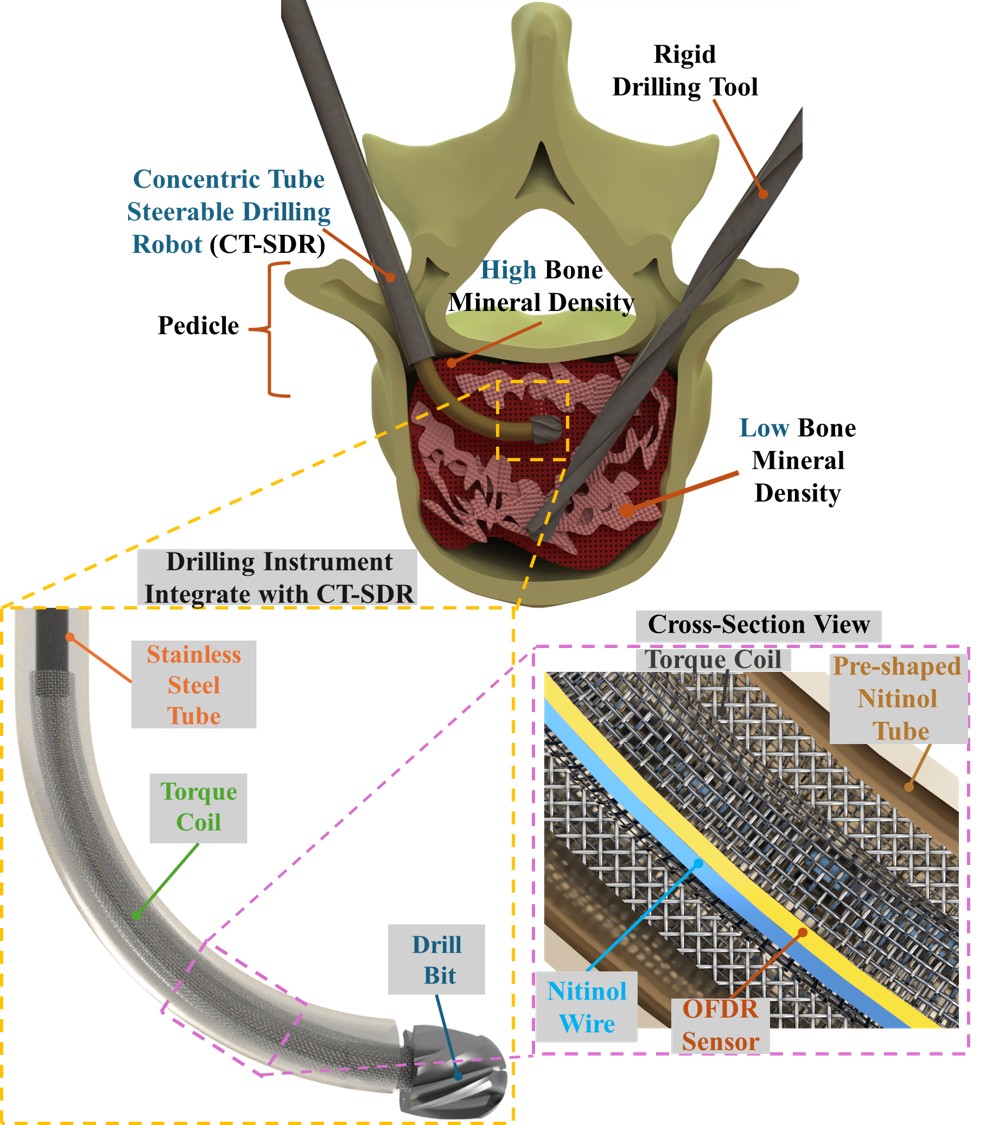}
     \caption{Conceptual illustration of the Sensorized CT-SDR compared to a conventional rigid drilling tool. Zoomed in view highlights the individual components of the CT-SDR. A cross-sectional view shows  the proposed OFDR-based SSA passing through  and housed within the internal channel of a flexible drilling instrument, which is guided by the pre-shaped NiTi tube of the CT-SDR.}
    \label{fig:Conceptual}
\end{figure}

Vertebral compression fractures are a common type of fracture in patients suffering from low bone mineral density \cite{Wright2014TheRP,Burge2007IncidenceAE,Johnell2006AnEO}. These fractures typically result in pain and spinal instability that, in severe cases, requires spinal fixation (SF) surgery. In this surgical procedure, as shown in Fig. \ref{fig:Conceptual}, a rigid drilling instrument is used to create a linear trajectory through pedicles of  multiple levels of vertebra. The drilled vertebrae are then conjoined together using rigid pedicle screws and rods. Despite this procedure being the gold standard for returning spinal mobility and relieving pain, it still suffers from pedicle screw fixation failure such as loosening and pullout \cite{weiser2017insufficient,wittenberg1991importance}. 

The aforementioned failure issues of SF can be tied to the complex anatomy of spine and the rigidity of current drilling instruments. The non-dexterous nature of these instruments forces surgeons to drill within a limited access corridor inside pedicle, preventing implants  to be fixated in an optimal region\cite{goldstein2015surgical}. Towards addressing these limitations, researchers have started exploring various continuum manipulators capable of creating curved drilling trajectories. Various tendon-driven continuum manipulators have been developed towards resolving the aforementioned issue. However, these systems still suffer from various limitations such as lacking structural stiffness for drilling \cite{alambeigi2017curved,alambeigi2019use} along with unintuitive manipulation due to the complex algorithms and sensors required for usage \cite{Ma2021AnAS}. To this end, researchers have recently introduced and validated a Concentric Tube Steerable Drilling Robot (CT-SDR) that is both structurally stiff and intuitive to utilize \cite{Sharma2023TBA,Sharma2023ACT,maroufi2025s3d,Kulkarni2025TowardsDD}. 

While the work of these researchers have been vital for overcoming the limitations of rigid drilling instruments, a new challenge arises from the utilization of the CT-SDR for SF. Unlike conventional rigid drilling systems that follow a straight, linear path, the CT-SDR creates a curved, nonlinear trajectory as shown in Fig. \ref{fig:Conceptual}. This complex path necessitates tracking of the robot's shape to ensure accuracy and safety during the drilling and fixation procedure. While imaging modalities such as optical trackers, X-ray, and ultrasound have been traditionally used for the guidance and navigation during SF procedure, each system suffers from individual shortcomings. For instance, optical trackers need a line-of-sight for tracking \cite{Sefati2018FBGBC}, X-ray introduces clinician to large amount of exposure to radiation over time \cite{Park2022RadiationSF}, and ultrasound systems can suffer from limited resolution \cite{Ryu2014FBGBSS}.

 Fiber Bragg Grating (FBG) optical shape sensors have recently gained popularity for their ability to overcome the aforementioned challenges while enabling seamless integration with both \textit{tendon-driven continuum manipulators} \cite{Sefati2021SurgicalRS,Sefati2016FBGbased,Sefati2019FBGBP} and \textit{concentric tube robots (CTRs) }\cite{Xu2016ShapeSF,Xu2016CurvatureTFS}. For instance, Sefati et al. \cite{Sefati2021SurgicalRS} proposed a surgical robotic system for treating pelvic osteolysis that incorporated two FBG shape sensors embedded within the wall thickness of a tendon-driven continuum manipulator, which was used for steerable drilling and milling of hard tissue. In this system, the FBG sensors enabled real-time tracking of the manipulator tip during bone cutting in a synthetic phantom. While effective, fabricating this sensing assembly is both complex and costly, as it requires precise laser engraving of Nitinol (NiTi) wires with small diameters to accommodate the placement of optical fibers.
In a separate study aimed at enabling shape sensing in concentric tube robots (CTRs), Xu et al. \cite{Xu2016CurvatureTFS,Xu2016ShapeSF} proposed a helically wrapped FBG-based sensing method capable of simultaneously measuring curvature and torsion. This was accomplished by machining helical grooves into the outer surface of a precurved NiTi tube using a custom-designed lathe to house three FBG sensors. Despite promising results, this approach presents significant fabrication challenges. Specifically, engraving grooves into thin-walled tubes is not only technically demanding but also increases the risk of structural failure due to elevated friction and strain during insertion, particularly when one tube is passed through another with minimal clearance \cite{Sefati2019FBGBP,Monet2020HighROF,Tavangarifard2024SingleFiberOF}.
Moreover, the direct exposure of helically-wrapped embedded sensors to bone during drilling raises additional concerns. Their interaction with the environment can increase the likelihood of sensor damage or tube failure, making this approach less suitable for CT-SDRs.

Aside from the fabrication challenges and the difficulty of integrating FBG sensors within the wall thickness or along the surface of pre-curved tubes in Concentric Tube Steerable Drills (CT-SDRs), FBG-based sensing also generally suffers from inherent limitations in spatial resolution. Specifically, its discrete strain measurement capability is constrained by the finite number of grating nodes along the FBG fiber \cite{Monet2020HighROF,Tavangarifard2024SingleFiberOF,Hanley2023OnTT}. This sparse distribution of sensing points can significantly degrade shape sensing accuracy, particularly during interactions with the environment, such as drilling through hard tissue \cite{Alambeigi2020SCADE}.

To collectively address the aforementioned challenges related to sensor fabrication, integration, and shape sensing in CT-SDRs, this paper introduces a novel shape sensing approach based on Optical Frequency Domain Reflectometry (OFDR), as our main contribution. Unlike FBGs, OFDR enables continuous strain measurement along the entire length of the fiber with enhanced spatial resolution \cite{nguyen2022toward, Tavangarifard2024SingleFiberOF}.
In the proposed method, a Shape Sensing Assembly (SSA) is first fabricated by integrating a single OFDR fiber with a flat NiTi wire. Following sensor calibration, and in contrast to previous helically wrapped, surface-mounted integration techniques in CTRs, the SSA is routed through and housed within the internal channel of the flexible drilling instrument, which itself passes through the pre-shaped NiTi tube of a cannulated CT-SDR. As shown in Fig. \ref{fig:Conceptual}, in this integration procedure, the drilling instrument  serves as a protective sheath for the SSA during the drilling process and is not integrated/adhered with the surface.
The performance of the proposed SSA integrated within a unique cannulated CT-SDR and its shape sensing capability was evaluated both in free-bending conditions and during drilling along multiple J-shaped trajectories inside synthetic bone phantoms.
\begin{figure}[t] 
    \centering
    \includegraphics[width=0.9\linewidth]{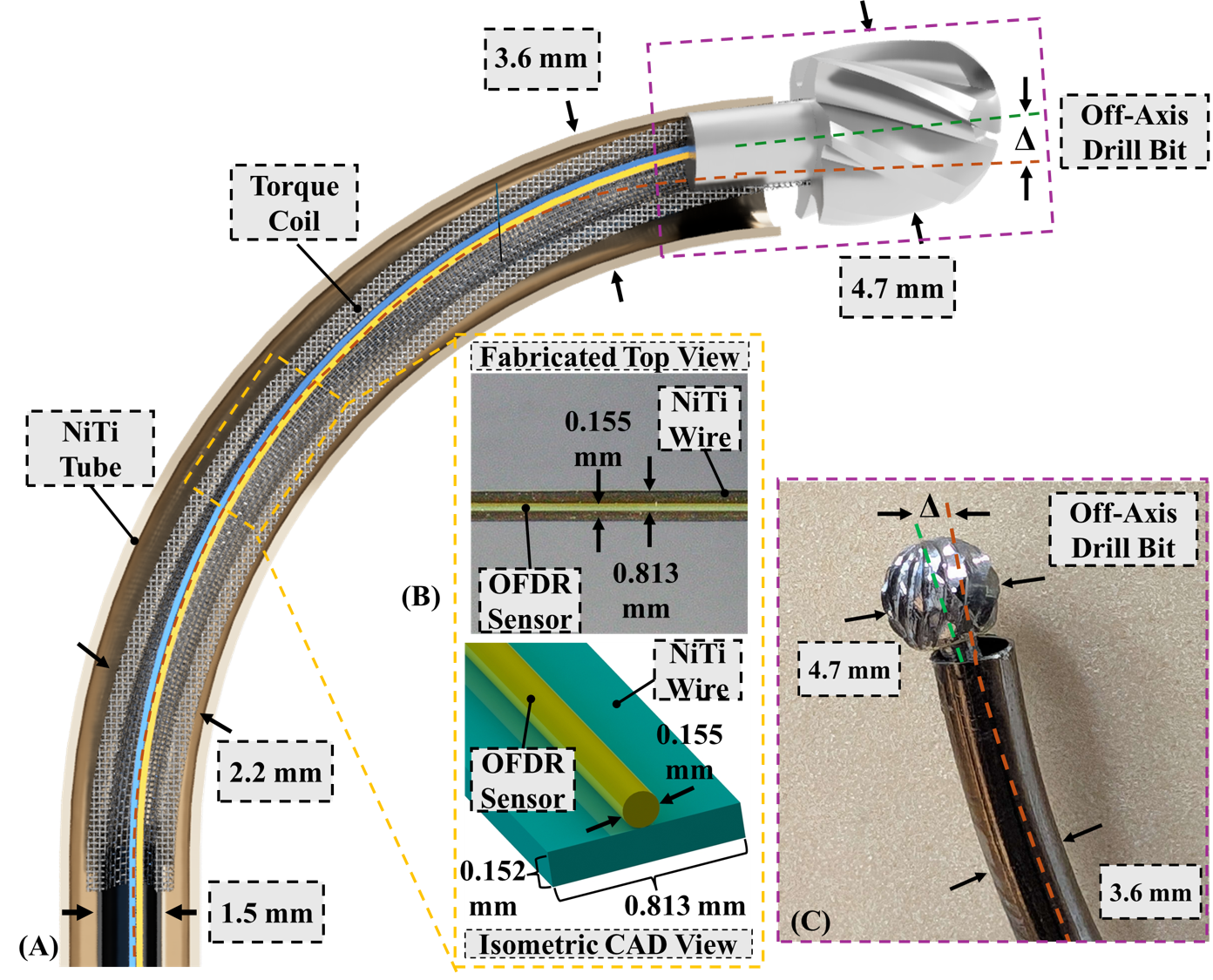}
     \caption{Rendered image of the Sensorized CT-SDR with labeled dimensions of its individual components. A zoomed-in view highlights the OFDR SSA and its dimensions in both top view and isometric view. The off-axis drill bit phenomenon is illustrated in the rendered image and further supported by an actual photograph of the drill bit, which also includes dimensional annotations.}
    \label{fig:Drill}
\end{figure}

\section{Shape Sensing Assembly and Steerable Drilling Robotic System} 
Unlike linear drills, the CT-SDR produces a curved trajectory inside the vertebral body, making it essential for clinicians to precisely monitor the drill’s path to avoid unintended damage. Precise shape information enables confirmation that the drill’s radius of curvature matches the intended trajectory. To achieve this, two requirements must be met: (i) the OFDR-based SSA  must be properly assembled and calibrated to ensure reliable and accurate strain measurement, and (ii) the CT-SDR must be designed to safely house the SSA while preserving its drilling capability in synthetic bone phantoms. Towards meeting these requirement, the following sections describe the proposed OFDR SSA, a unique cannulated CT-SDR, and their integration in greater detail.

\subsection{Shape Sensing Assembly Fabrication and Calibration}
To fabricate and calibrate the SSA, we followed the procedure outlined in \cite{Tavangarifard2024SingleFiberOF,nguyen2022toward}. This process consists of the following four critical steps: 

\textbf{(Step I)} The process starts with  determining the proper substrate dimensions for securing the OFDR fiber. Consistent with the aforementioned studies and as opposed to most of the FBG-based and OFDR-based SSAs (e.g., \cite{Sefati2021SurgicalRS,Xu2016ShapeSF,Monet2020HighROF}), we simplified  fabrication by using a single OFDR fiber attached to a flat rectangular NiTi wire substrate. To select the correct flat rectangular NiTi wire, we first determined the neutral plane of the SSA  based on the Young's Modulus and geometry of the OFDR fiber and NiTi wire. Considering the 0.155 mm diameter of the OFDR fiber (HD65, Luna Innovations Inc.) and using the optimization procedure described in \cite{nguyen2022toward}, the most optimal wire dimension of 0.152 mm $\times$ 0.813 mm was obtained. This dimension is essential because it directly affects the maximum bending radius, induced strain, and overall sensor sensitivity. The dimensions of the designed OFDR sensor and selected NiTi wire  are shown in Fig. \ref{fig:Drill}.

\begin{figure}[t] 
    \centering
    \includegraphics[width=1\linewidth]{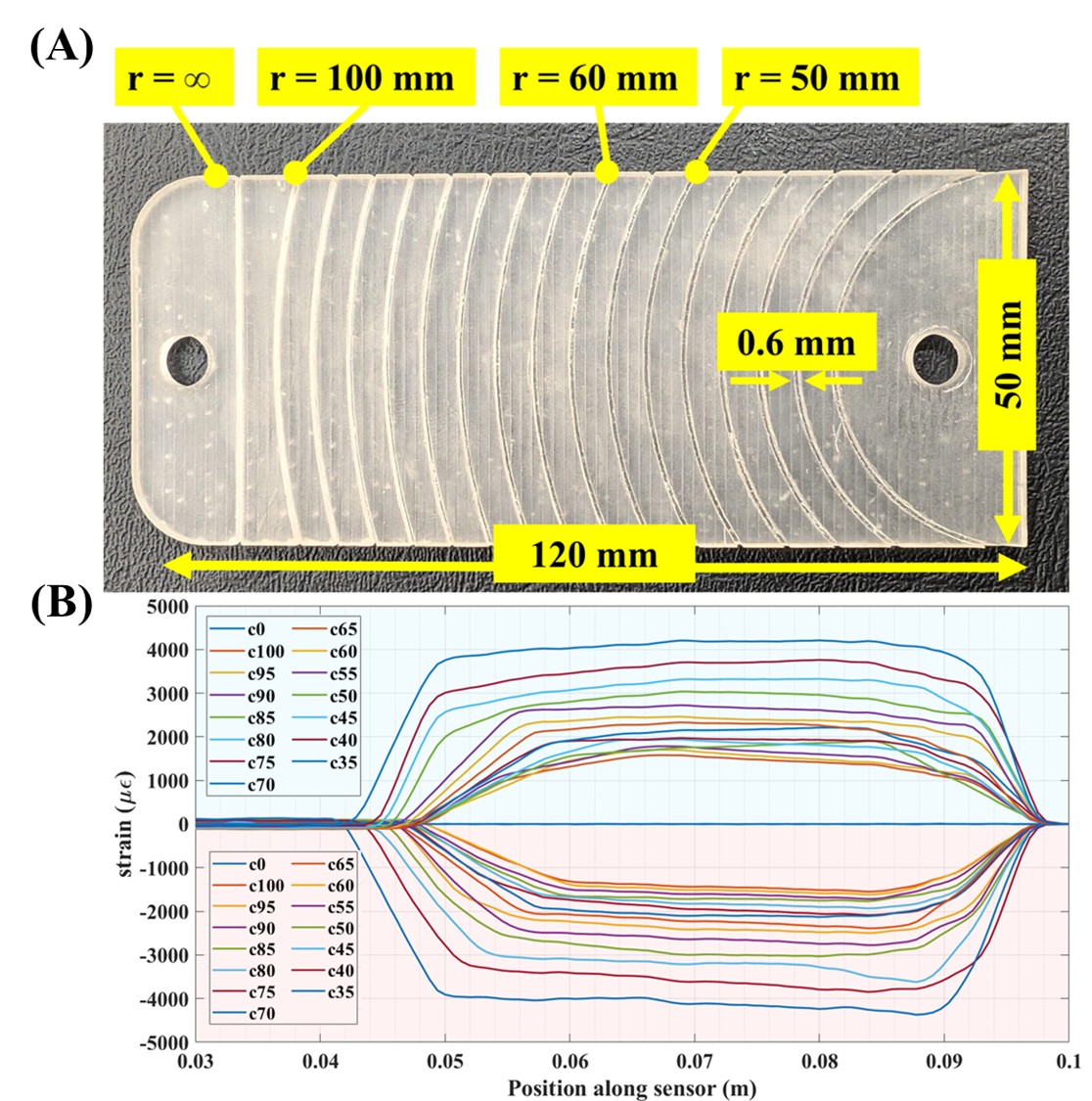}
     \caption{(A) Calibration jig with slots of varying radii of curvature. (B) Measured strain data collected in both positive and negative direction by OFDR SSA for various radii of curvature based on custom jig measurement.}
    \label{fig:JigStrain}
\end{figure}

\textbf{(Step II)} After selecting the optimal wire dimension, the second critical step  focused on attaching the OFDR sensor to the Niti Wire. After cleaning the surface of wire and fixating it to a flat surface, the OFDR fiber was  aligned along the wire's centerline  and bonded using cyanoacrylate adhesive (Loctite 495)  over a length of 30 cm. After letting the assembly cure for approximately 24 hours, the OFDR SSA was ready for use. 

\begin{figure}[t] 
    \centering
    \includegraphics[width=0.8\linewidth]{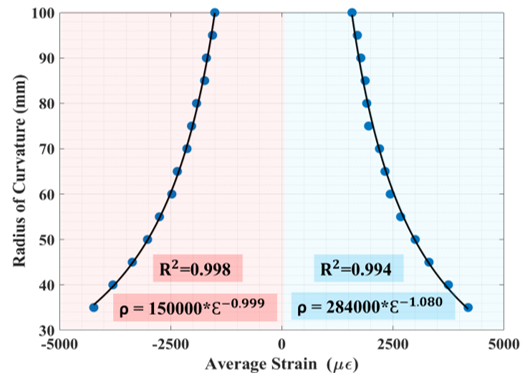}
     \caption{Calibration between strain and radius of curvature for the SSA in both positive and negative bending directions. }
    \label{fig:calibration}
\end{figure}

\textbf{(Step III)} The final step   focused on calibrating the sensor to establish a mathematical relationship between the measured strain and radius of curvature of SSA. This step is essential to ensures reliable shape reconstruction from the strain data collected with the  SSA. To this end, as shown in Fig. \ref{fig:JigStrain}-A, a calibration jig was first designed and fabricated with Clear resin using an SLA 3D printer (Form3, Formlabs Inc) \cite{nguyen2022toward,Tavangarifard2024SingleFiberOF} and strain data was collected by placing the SSA in a linear slot as well as slots with radii of curvatures ranging from 35 mm to 100  mm, 0.6 mm slot width, in 5 mm increments. Of note, data was recorded using the OFDR interrogator (ODiSI 6000 Series, Luna Innovations Inc.) at a frequency of 31.25 Hz and a 0.65 mm spatial resolution. To capture both tensile and compressive responses, calibration measurements were performed in both positive and negative directions with three trials conducted per curvature. Figure \ref{fig:JigStrain}-B shows the collected strain during the calibration procedure for both positive $\varepsilon_{SSA,positive}$ and negative $\varepsilon_{SSA,negative}$ strains. The following equations shows the fitted curve  representing the strain-curvature relationships for the positive and negative directions of the SSA:
\begin{align}
\rho_{\text{SSA,positive}} &= 284000 \, \cdot \varepsilon_{\text{SSA,positive}}^{-1.08} \\
\rho_{\text{SSA,negative}} &= 150000 \, \cdot \varepsilon_{\text{SSA,negative}}^{-0.999}
\end{align}

where $\rho_{\text{SSA,positive}}$ and $\rho_{\text{SSA,negative}}$ are the bending radii of curvature (mm) for both positive and negative strain values, respectively. $\epsilon$ is the measured strain in both positive and negative direction. ($\mu\varepsilon$). Figure \ref{fig:calibration} shows the results of this calibration procedure.

\textbf{(Step IV)} Using the obtained calibrated strain–curvature relationship, the 2D shape of the SSA can be reconstructed. For each arc length increment $\Delta s$, the corresponding curvature $\kappa$, radius of curvature $\rho$, and local orientation $\theta(s)$ are computed as:
\begin{equation}
\kappa = \frac{1}{\rho}, \qquad \Delta \theta = \kappa , \Delta s.
\end{equation}
The Cartesian displacements $\Delta x$ and $\Delta y$ are then given by:
\begin{equation}
\Delta x = \cos \theta(s),\Delta s, \qquad \Delta y = \sin \theta(s),\Delta s.
\end{equation}

\begin{figure*}[t] 
    \centering
    \includegraphics[width=0.91\linewidth]{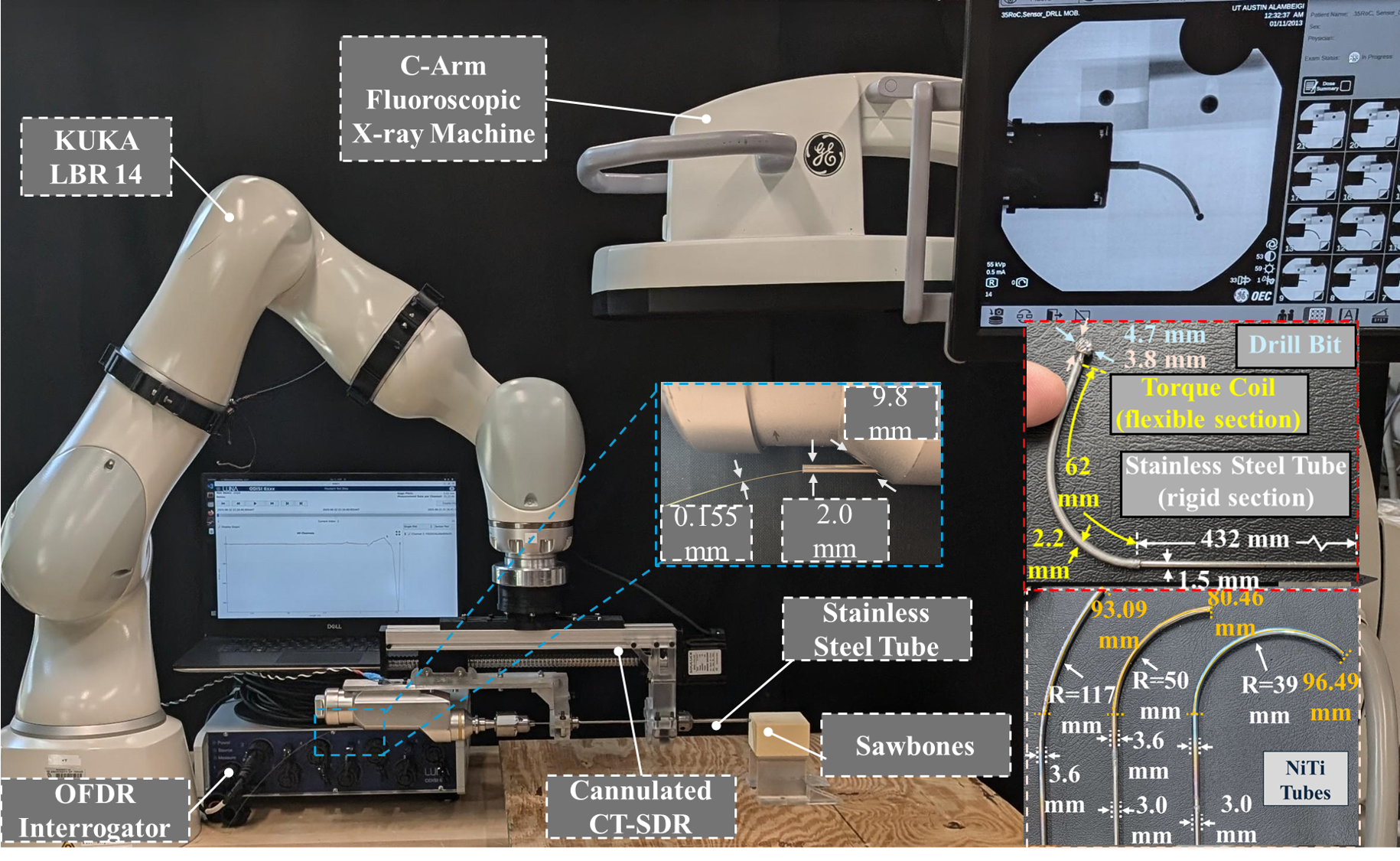}
     \caption{Experimental Setup consisting of a KUKA Robot arm, Fluoroscopic X-Ray machine, Cannulated CT-SDR, and the OFDR shape sensor and interrogator. Close up views of the sensor inside the flexible drilling instrument is shown along with a zoomed in view of the flexible drilling instrument with dimensions and the NiTi tubes.}
    \label{fig:ExperimentalSetUp}
\end{figure*}

By integrating these increments over the arc length, the full shape of the SSA can be reconstructed.
\subsection{A Cannulated Concentric Tube Steerable Drilling Robot }
Inspired by the CT-SDR design proposed in \cite{Sharma2023ACT},  we introduce a \textit{cannulated CT-SDR} that enables safe passage and secure housing of the SSA from behind through the flexible drilling instrument, where the instrument itself functions as the guiding and protective sheath. As illustrated in Fig. \ref{fig:ExperimentalSetUp}, the two degree-of-freedom (DoF) cannulated CT-SDR comprises three main components: 

(i) J-shaped, heat-treated NiTi tubes (Euroflex GmbH, Germany) with 3.6 mm outer diameter (OD), 3.1 mm internal diameter (ID), and length that is constrained within a straight stainless steel tube (McMaster-Carr) with OD = 5 mm , ID = 4 mm, and length of 85 mm (shown in Fig. \ref{fig:ExperimentalSetUp}). The NiTi tube geometry is determined by clinically pre-planned drilling trajectories and serves to guide the flexible drilling instrument. 

(ii)  A two-DoF instrument rotation and tube insertion actuation mechanism (Fig. \ref{fig:ExperimentalSetUp}). Rotational power for the flexible drilling instrument is provided by a modified off-the-shelf handheld surgical drill (256919266906, eBay). As shown in the zoomed-in view of  Fig. \ref{fig:ExperimentalSetUp}, the cannulated configuration of this drill allows the SSA to pass through the hollow flexible instrument from the back of the drill, representing a critical upgrade over previous non-cannulated CT-SDRs \cite{Sharma2023ACT,sharma2024patient}. Linear actuation  is used to push the constrained pre-curved NiTi tubes outside the straight stainless still tube and is provided by a stage driven by a NEMA 23 stepper motor (B08ZKF9ZD8, Amazon) controlled via an Rtelligent R86 driver (B07S897BZZ, Amazon). An Arduino Nano generates step and direction signals, while a joystick provides user input, with stage travel set to 10 mm per command and a resolution of 0.030 mm.

\begin{figure*}[t] 
    \centering
    \includegraphics[width=0.88\linewidth]{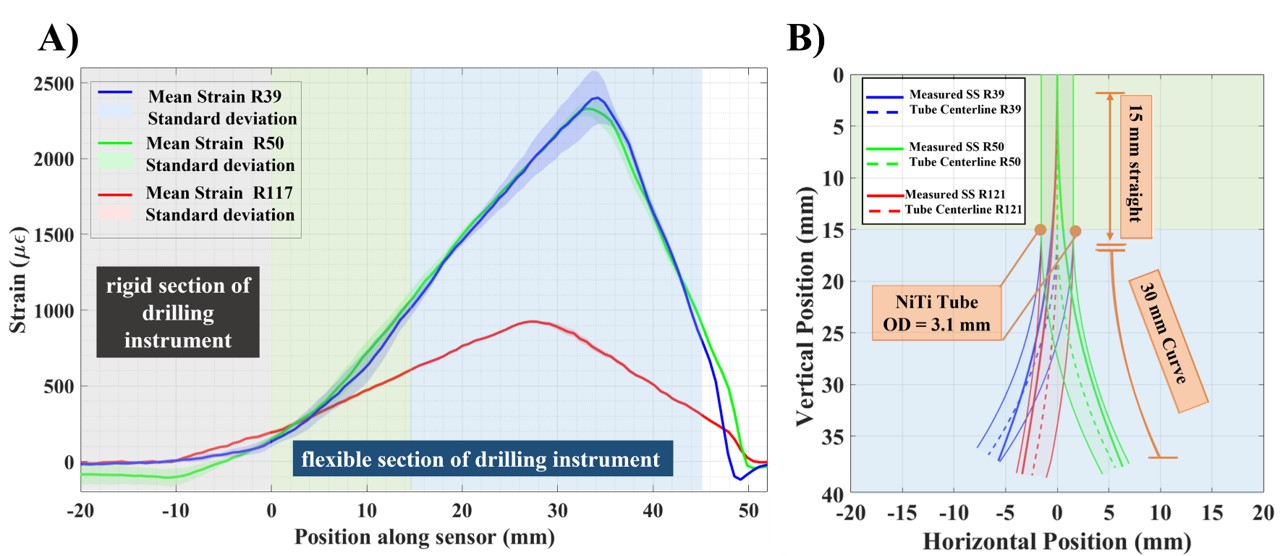}
     \caption{(A) Strain along the length of the OFDR SSA during the Free Bending experiments for three radii of curvature. Shaded regions indicating the rigid section and the flexible section of the flexible drilling instrument. Standard deviations are also shown around the strain curves. (B) Reconstructed shape from measured strain across straight and curved section.}
    \label{fig:FBStrainShape}
\end{figure*}

(iii) As shown in Fig. \ref{fig:ExperimentalSetUp}, a custom-made, cannulated flexible drilling instrument, consisting of a 4.7 mm OD and 3.8 mm length machined drill bit (8175A74, McMaster-Carr) laser-welded to a torque coil (Asahi Intecc, USA, Inc.) with 2.2 mm OD and  62 mm length. To extend its length and ensure structural integrity during drilling, a 1.5 mm OD and 120 mm length stainless-steel tube (51755K13, McMaster-Carr) was laser-welded to the torque coil. The drilling instrument has the overall length of  500 mm with a maximum OD of 2.0 mm and can be passed through the NiTi tubes with a maximum inner diameter (ID) of 3.1 mm. As shown in Fig. \ref{fig:Drill}, unlike previous approaches that embedded the SSA within internal channels in the robot’s wall thickness \cite{Sefati2016FBGbased} or machined grooves on the surface of CTR tubes \cite{Xu2016ShapeSF}, the hollow structure of the flexible drilling instrument is instead utilized as a sheath to safely guide and protect the SSA during the drilling process.
Figure \ref{fig:Drill} shows the integration of the drilling instrument with a pre-curved NiTi tube.

\section{Experimental Evaluation and Results}
To thoroughly evaluate the performance of the proposed SSA and its integration within the cannulated flexible drilling instrument, we first mounted the cannulated CT-SDR onto a seven-DoF robotic manipulator (KUKA LBR Med 14, Germany), as shown in Fig. \ref{fig:ExperimentalSetUp} \cite{Sharma2024BiomechanicsARA}. The SSA was then fully inserted through the cannulated channel of the CT-SDR into the flexible drilling instrument. To prevent changes in sensing length or accidental displacement during drilling, the OFDR fiber at the CT-SDR entry point was fixated to the CT-SDR body. The opposite end of the fiber was connected to an interrogator (ODiSI 6000 Series, Luna Innovations Inc.) for dynamic strain measurement. During all experiments, data were acquired at a frequency of 31.25 Hz with a spatial resolution of 0.65 mm.
We conducted two sets of experiments using this setup to validate the SSA and its proposed integration with CT-SDR: (i) free-bending tests, and (ii) drilling experiments inside synthetic  Sawbones (Pacific Research Laboratories, USA) phantoms  along four trajectories (i.e., one straight path and three J-shaped paths) using pre-shaped NiTi tubes with radii of curvature of 39 mm, 50 mm, and 117 mm listed  in Fig. \ref{fig:ExperimentalSetUp}.

\subsection{Free-Space Bending Test}
To establish a baseline behavior of the Sensorized CT-SDR system, free bending test were conducted without drilling into synthetic bone samples. Without loss of generality, we assumed a surgeon-selected trajectory consisting of a 15 mm straight insertion followed by a 50 mm curved trajectory. These dimensions approximately match an L3 vertebra \cite{Zindrick1986ABS}. 
For the straight insertion, the procedure consisted of: (i) accelerating the drill to 320 rpm, (ii) inserting 10 mm at 1.5 mm/s using the robotic arm, (iii) decelerating to 0 rpm, and (iv) recording strain data with the LUNA Interrogator. This sequence was repeated twice to complete the 15 mm straight trajectory. Afterward, the NiTi tube was advanced along a curved trajectory at 1.5 mm/s for 50 mm, with the same cycle of acceleration, deceleration, and data collection performed at 10 mm increments. The entire procedure was repeated in reverse and performed three times for each of the  considered three J-shape  trajectories. Figure \ref{fig:FBStrainShape} illustrates the results of these experiments corresponding to the measured strain and reconstructed shape of the drill, respectively.

\subsection{Drilling Experiments in Sawbones Phantoms}
\begin{figure*}[t] 
    \centering
    \includegraphics[width=0.9\linewidth]{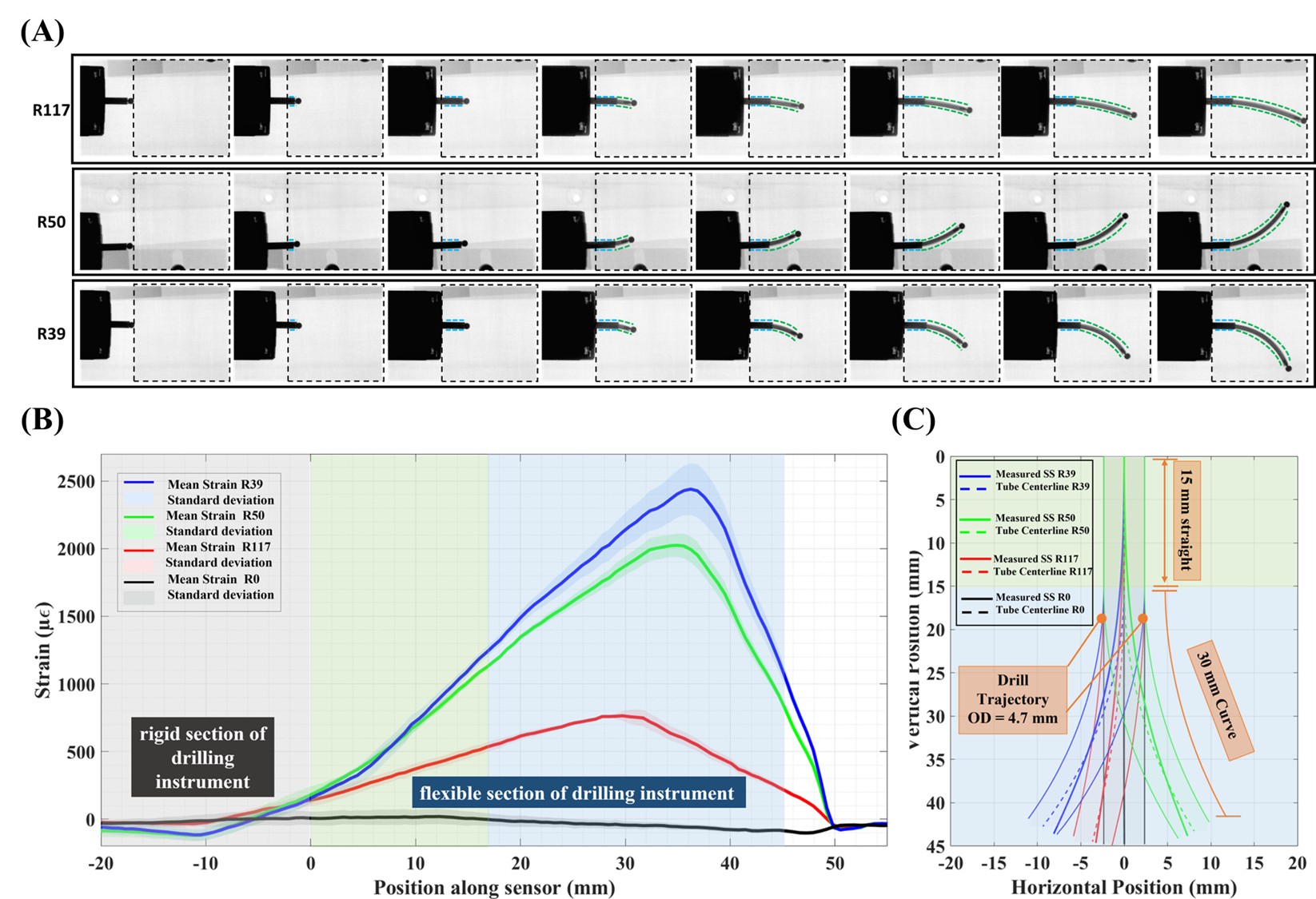}
     \caption{(A) Fluoroscopic images taken during the drilling procedure showing the CT-SDR with OFDR sensor drilling different radii of curvatures. The drilled trajectory is shown in blue dashed lines for the linear trajectories and in orange dashed lines for the curved trajectory. The Sawbones border is marked in black dashed lines. (B) Strain along the length of the OFDR SSA during the drilling experiments and (C) shape reconstruction comparing measured and expected drilled trajectories from drilling experiments. }
    \label{fig:DrillingStrainShape}
\end{figure*}

In this set of experiments, drilling was performed inside PCF 5 Sawbones phantoms (Pacific Research Laboratories, USA). As shown in Fig. \ref{fig:ExperimentalSetUp}, each phantom was first securely fixated to the table, after which the sensorized CT-SDR mounted on the robotic manipulator was autonomously aligned with the starting position of the phantom. The drill’s rotational speed was then accelerated to 320 RPM, and the same experimental protocol as the free-bending tests was applied.
During drilling, a C-Arm fluoroscopic X-ray machine (OEC One CFD, GE Healthcare) was used to acquire images at every 10 mm increment, enabling tracking of the CT-SDR shape  during the experiments. Each of the four trajectories (one straight and three J-shaped) was repeated three times using NiTi tubes shown in Fig. \ref{fig:ExperimentalSetUp}. After drilling, the Sawbones samples were halved, and the resulting trajectories were evaluated in SolidWorks (Dassault Systèmes) to measure their radii of curvature, which were used as ground truth for evaluating the SSA’s shape-sensing accuracy.
Figure \ref{fig:DrillingStrainShape} presents representative X-ray images alongside the measured strain and reconstructed shapes obtained from these experiments.

\subsection{Evaluation Metrics}
Two evaluation metrics were used to assess the reconstruction accuracy of the OFDR-based SSA integrated with the cannulated CT-SDR during both free-space bending and drilling experiments: \textit{Average Tip Error} and \textit{Average Shape Error}. The \textit{Average Tip Error} was defined as the L2 norm of the difference between the reconstructed tip position and the expected tip position of the flexible drilling instrument (reported as an absolute error in millimeters): 
\begin{equation}
E_{\text{Tip Error}} \;=\; 10^{3}\,
\left\lVert 
\begin{bmatrix} x^{\text{em}}_{L} \\[2pt] y^{\text{em}}_{L} \end{bmatrix}
-
\begin{bmatrix} x^{\text{gt}}_{L} \\[2pt] y^{\text{gt}}_{L} \end{bmatrix}
\right\rVert_{2}
\;\;\text{[mm]}
\end{equation}

The \textit{Average Shape Error} was computed as the mean L2 norm of the differences between reconstructed points along the instrument and their actual positions (also reported as an absolute error in millimeters):
\begin{equation}
E_{\text{Shape Error}} \;=\; 10^{3}\,\frac{1}{L}\sum_{i=1}^{L}
\left\lVert 
\begin{bmatrix} x^{\text{em}}_{i} \\[2pt] y^{\text{em}}_{i} \end{bmatrix}
-
\begin{bmatrix} x^{\text{gt}}_{i} \\[2pt] y^{\text{gt}}_{i} \end{bmatrix}
\right\rVert_{2}
\;\;\text{[mm]}
\end{equation}

where, $(x^{\text{em}}_{i}, y^{\text{em}}_{i})$ are the measured  coordinates reconstructed by the OFDR-based shape-sensing assembly, and $(x^{\text{gt}}_{i}, y^{\text{gt}}_{i})$ are the corresponding ground-truth centerline coordinates. The index $i$ runs over $L$ samples along the 45\,mm instrument length, with data acquired at a gauge pitch of 0.65\,mm (one measurement approximately every 0.65\,mm). The results of these metrics for all experiments are summarized in Table \ref{tab:errors_table}. 
Reported values correspond to the {absolute mean errors computed across the three repeated 
trials for each experimental condition. The percentages shown in brackets indicate the 
normalized errors with respect to the corresponding instrument length.

\section{Discussion}
The results of the free-space bending test are presented in Fig. \ref{fig:FBStrainShape}-A (strain profiles), Fig. \ref{fig:FBStrainShape}-B (shape reconstructions), and Table \ref{tab:errors_table} (numerical metrics). In Fig. \ref{fig:FBStrainShape}-A, negative x-axis values correspond to the portion of the SSA located within the rigid, straight section of the flexible drilling instrument, where negligible strain was measured. Positive values, in contrast, indicate SSA bending within the flexible section of the instrument guided by the J-shaped trajectory. As expected, the measured strain increased linearly as the robot followed the J-shaped trajectory. This pattern was consistent across all NiTi tubes, with the maximum strain magnitude increasing as the tube's radius of curvature decreased (e.g., 2400 $\mu\varepsilon$ for R39 vs. 900 $\mu\varepsilon$ for R117).
It is worth noting that, instead of exhibiting a near-zero strain followed by a sharp increase at the transition into the curved section, the SSA measurements showed a more gradual linear increase. As shown in Fig. \ref{fig:Drill}, this behavior can be attributed to the loose fit of the SSA inside the flexible drilling instrument and the diameter mismatch between the instrument and the pre-curved NiTi tube. These size gaps reduced the sensitivity of the sensor in rapidly registering local strain and curvature changes. Compared with other integration strategies (e.g., embedding within the wall thickness or surface grooves), our approach is mechanically simpler but comes at the cost of reduced strain sensitivity. 
Despite this limitation, the reconstructed shapes (Fig. \ref{fig:FBStrainShape}-B) and the evaluation metrics (Table \ref{tab:errors_table}) confirm high reconstruction accuracy, with a maximum average tip error of 1.32 mm for the R39 tube and an overall average normalized shape error of less than 1.5\%. Notably, this error is lower than the 6\% error reported in \cite{Monet2020HighROF} for a tendon-driven steerable drilling robot (maximum length 35 mm) that required two SSAs, each using three FBG fibers within tightly machined  channels within the wall thickness of robot, or alternatively, two OFDR SSAs embedded within the mentioned channels. In contrast, our approach relies on a single OFDR fiber attached to a flat wire, which not only simplifies and reduces fabrication cost but also allows the assembly to be loosely inserted into the instrument channel without requiring custom grooves or channels. These results clearly demonstrate the strong shape reconstruction capability of the proposed SSA and its integration approach.

The results of the drilling experiments in Sawbones phantoms are presented in Fig. \ref{fig:DrillingStrainShape} and Table \ref{tab:errors_table}. Unlike free-bending tests, drilling introduces additional forces, vibrations, and tool–bone interactions that can challenge the sensing accuracy of the SSA. Nevertheless, the results demonstrate that the proposed SSA and its integration remained robust against these external disturbances. The X-ray snapshots in Fig. \ref{fig:DrillingStrainShape}-A confirm that the Sensorized CT-SDR can safely house the OFDR-based SSA during drilling while still enabling reliable data collection. 
As shown in Fig. \ref{fig:DrillingStrainShape}-B, the strain patterns for J-shaped drilling trajectories were consistent with expectations: larger radii of curvature produced lower strain values, while straight trajectories produced near-zero strain. Similar to the free-bending experiments, the mismatch between the diameters of the tube, drilling instrument, and SSA reduced sensor sensitivity, leading to a gradual linear increase in measured strain rather than the sharp rise expected at the onset of curvature. Despite this limitation, the reconstructed shapes remained accurate, with a maximum average tip error of 1.73 mm and an average shape error of 0.44 mm. Table \ref{tab:errors_table} and Fig. \ref{fig:DrillingStrainShape}-C highlight the errors seen during the drilling tests. R0 was used for baseline testing and demonstrated a low average tip error of 0.20 mm (0.44\%) and an average shape error of 0.07 mm (0.16\%). For the curved drilling tests, R46 exhibited the highest errors with 1.73 mm (3.84\%) tip error and 0.44 mm (0.98\%) shape error due to the difficulty of following the tight curvature under drilling forces. R121 had the lowest errors at 0.46 mm (1.02\%) and 0.22 mm (0.49\%), consistent with the easier geometry of a gentle bend. R53 was intermediate, with a tip error of 1.12 mm (2.48\%) and shape error of 0.38 mm (0.85\%).  

An investigation of the shape reconstruction results shown in Figs. \ref{fig:FBStrainShape} and \ref{fig:DrillingStrainShape} for both free-space bending and drilling experiments demonstrates that, in nearly all J-shaped trajectories, the reconstructed shapes exhibit larger radii of curvature compared to the expected midline trajectory. This error can be attributed to the elastic nature of the torque coil section of the drilling instrument, which tends to return toward its original straight configuration, thereby pulling the embedded SSA with it. As a result, the measured curvature is reduced, giving the appearance of a larger reconstructed radius of curvature. This explains why the reconstructed shape tends to align closer to the outer surface of the NiTi tube, following a path with higher radius of curvature rather than the true midline trajectory. This off-axis alignment of the drilling instrument relative to the tube centerline can be clearly observed in Fig. \ref{fig:Drill}-C, where the instrument is shown inside the NiTi tube. Nevertheless, despite this misalignment, the reconstructed shape errors remain reasonable for a steerable drilling robot and are better than previously reported results in the literature (e.g., \cite{Monet2020HighROF}).

Overall, the results from both free bending and drilling confirm that the OFDR SSA maintains reliable accuracy in controlled and clinically relevant conditions. Importantly, all errors remained below 2 mm, which is within the clinically acceptable margin for spinal procedures where screw misplacements of less than 2 mm are generally considered safe and not associated with complications \cite{winder2017accuracy}. This demonstrates that the proposed integration achieves clinically relevant accuracy while maintaining performance under both controlled and realistic drilling conditions.

\begin{table}[t]
\centering
\caption{Tip and shape errors for free-space bending and drilling experiments}
\normalsize
\renewcommand{\arraystretch}{1.25} 
\setlength{\tabcolsep}{6pt}
\begin{tabular}{ccc}
\toprule
\makecell{\textbf{Radius of} \\ \textbf{Curvature} \\ (mm)} & \makecell{\textbf{Average Tip} \\ \textbf{Error} \\ (mm)} & \makecell{\textbf{Average Shape} \\ \textbf{Error} \\ (mm)} \\
\midrule
\multicolumn{3}{l}{\textbf{w/o Sawbones}} \\

$\mathbf{R117}$ & 0.97 [2.14\,\%] & 0.59 [1.31\,\%]  \\
$\mathbf{R50}$  & 0.78 [1.73\,\%] & 0.67 [1.49\,\%]  \\
$\mathbf{R39}$  & 1.32 [2.92\,\%] & 0.41 [0.92\,\%]  \\

\midrule
\multicolumn{3}{l}{\textbf{w/ Sawbones}} \\
$\mathbf{R_{\infty}}$ & 0.20 [0.44\,\%] & 0.07 [0.16\,\%]  \\
$\mathbf{R121}$       & 0.46 [1.02\,\%] & 0.22 [0.49\,\%]  \\
$\mathbf{R53}$        & 1.12 [2.48\,\%] & 0.38 [0.85\,\%]  \\
$\mathbf{R46}$        & 1.73 [3.84\,\%] & 0.44 [0.98\,\%]  \\
\bottomrule
\end{tabular}
\label{tab:errors_table}
\end{table}

\section{Conclusion and Future Work}
To advance safe spinal fixation surgery using CT-SDRs, this paper introduced a method for integrating and utilizing an OFDR sensor within the flexible drilling instrument of a cannulated CT-SDR, enabling accurate strain measurement and shape reconstruction during both free-bending and drilling in synthetic bone phantoms. To the best of our knowledge, this is the first demonstration of a CT-SDR with a cannulated drill capable of safely housing an OFDR-based SSA within its flexible drilling instrument. The proposed design was validated through static testing and drilling experiments, showing reliable strain measurements and accurate shape reconstruction with average tip and shape errors of less than 1.8 mm and 0.7 mm, respectively.
In future work, we will evaluate the efficacy of the drill \cite{Maroufi2026SystematicCDP} and the integration and SSA design through drilling experiments in more realistic phantom bones, animal bones, and human cadaveric specimens along with flexible implants \cite{Kulkarni2024SFF,Kulkarni2025SynergisticPSA, Kulkarni2025DIOFDR,kulkarni2026CurvedSIJF,sharma2024spatial,Kulkarni2025ABSF}.

\bibliographystyle{./IEEEtran}
\bibliography{./root}

\end{document}